\documentclass[11pt]{article}

\usepackage{unicode-math}

\usepackage[a4paper,margin=1in]{geometry}
\usepackage{amsmath,amsthm}
\usepackage{microtype}
\usepackage{graphicx}
\usepackage{booktabs}
\usepackage{multirow}
\usepackage{array}
\usepackage{tabularx}
\usepackage{caption}
\usepackage{subcaption}
\usepackage{float}
\usepackage{placeins}

\usepackage{xurl}
\usepackage[hidelinks]{hyperref}

\newtheorem{definition}{Definition}
\newtheorem{proposition}{Proposition}

\title{Learning--Inference Concurrency in DynamicGate-MLP:\\
Structural and Mathematical Justification}
\author{Yongil Choi}
\date{2026. 2. 6}

\begin{document}
\maketitle
\begin{center}
    Sorynorydotcom Co., Ltd./AI Open Research Lab \href{mailto:hurstchoi@sorynory.com}{\nolinkurl{hurstchoi@sorynory.com}}

\url{https://orcid.org/0009-0009-8813-5420}
\end{center}

\begin{abstract}
\noindent
Conventional neural networks strictly separate learning and inference because if parameters are updated during inference, outputs become unstable and even the inference function itself is not well-defined~\cite{bottou2010sgd,shalevshwartz2012online,zinkevich2003online}. This paper shows that \textit{DynamicGate-MLP} structurally permits learning--inference concurrency~\cite{bengio2015conditional,shazeer2017moe}. The key idea is to separate \textbf{routing (gating) parameters} from \textbf{representation (prediction) parameters}, so that the gate can be adapted online while inference stability is preserved, or weights can be selectively updated only within the inactive subspace~\cite{bengio2015conditional,shazeer2017moe,evci2020rigl,han2015pruning}. We mathematically formalize sufficient conditions for concurrency and show that even under asynchronous or partial updates, the inference output at each time step can always be interpreted as a forward computation of a valid model snapshot~\cite{niu2011hogwild,li2014parameterserver,dean2012distbelief}. This suggests that DynamicGate-MLP can serve as a practical foundation for online-adaptive and on-device learning systems~\cite{gama2014conceptdrift,wang2021tent}.
\end{abstract}

\section{Introduction}
Modern neural networks are typically deployed under a \textit{train--then--infer} paradigm~\cite{bottou2010sgd,kingma2015adam}. When training begins, inference is stopped; when inference is performed, parameters are fixed~\cite{bottou2010sgd}. This separation is not merely a simple system design choice but \textbf{inevitable}, because in dense models, if parameters are continuously updated, the inference function becomes non-stationary and is not mathematically well-defined~\cite{rumelhart1986backprop,shalevshwartz2012online,zinkevich2003online}.

However, many real systems---such as edge AI, continual-learning agents, adaptive control, and personalized models---must \textbf{perform inference while learning at the same time}~\cite{parisi2019lifelong,delange2019survey}. The limitation that standard architectures cannot provide such concurrency calls for a new structural design~\cite{parisi2019lifelong,kirkpatrick2017ewc}.

DynamicGate-MLP activates only a subset of neurons/blocks for each input via input-dependent gating and conditional computation~\cite{bengio2015conditional,shazeer2017moe,fedus2022switch}. This structure implies not only reduced computation, but more fundamentally a crucial property that can make \textbf{learning--inference concurrency} mathematically valid~\cite{bengio2015conditional,shazeer2017moe}.

This paper formalizes why concurrency is possible in DynamicGate-MLP and presents sufficient conditions and proofs~\cite{shalevshwartz2012online,zinkevich2003online}.

\section{Related Work}
\label{sec:related_work}

Research on coping with distribution shifts (e.g., domain shift, drift) at inference time has largely evolved along two lines: \emph{Test-Time Adaptation (TTA)} and \emph{Test-Time Training (TTT)}.
These approaches typically assume a model trained offline and then deployed, but aim to recover performance by updating \emph{a subset of parameters} without labels (or with weak signals) at test time, or by gradually adapting in an input stream.
However, most existing methods adjust parameters while keeping the model's \emph{computational structure (active paths) fixed}; approaches that explicitly incorporate \emph{structural sparse execution}---i.e., reducing computation (FLOPs) in an input-dependent manner---are relatively rare.

\subsection{Normalization-Based Test-Time Adaptation}
\label{subsec:tta_norm}
The simplest form of partial adaptation relies on re-estimating normalization statistics (e.g., BatchNorm) under the target distribution.
AdaBN adapts to domain shifts by replacing BatchNorm statistics with those estimated from the target domain \cite{Li2016AdaBN}.
TENT performs test-time adaptation by minimizing predictive entropy, while restricting updates primarily to BatchNorm-related parameters (statistics and affine terms) to improve stability and efficiency \cite{wang2021tent}.
TTN points out that BatchNorm statistics can become unreliable under practical conditions such as small batch sizes and non-i.i.d. test streams, and proposes a stabilized adaptation strategy by interpolating between training-time and test-time statistics \cite{Lim2023TTN}.
While these normalization-based methods are easy to implement and clearly realize ``partial updates,'' they fundamentally keep the active computational paths fixed and focus on increasing prediction confidence rather than structurally controlling computational efficiency.

\subsection{Stabilizing Entropy-Minimization TTA}
\label{subsec:tta_stability}
With the widespread adoption of entropy-minimization TTA, many studies have sought to mitigate collapse, error accumulation, and forgetting during online adaptation.
EATA introduces selective updates by choosing informative samples and adding regularization to preserve important parameters, thereby reducing degradation in streaming settings \cite{Niu2022EATA}.
SAR aims to improve stability in ``wild'' environments by filtering unreliable samples and incorporating a sharpness-aware perspective into entropy-based adaptation \cite{Niu2023SAR}.
CoTTA addresses continual shifts in test streams by combining prediction averaging (e.g., augmentation/weight averaging) with partial restoration mechanisms to suppress long-term forgetting \cite{Wang2022CoTTA}.
Although these methods provide practical stabilization mechanisms for sustained test-time updates, they still assume a fixed computational graph and do not directly model or control \emph{structural} variability such as fluctuating routing decisions.

\subsection{Test-Time Training: Self-Supervised Sample/Stream Adaptation}
\label{subsec:ttt}
TTT methods convert test inputs into self-supervised learning tasks and perform short updates at the sample or stream level before making predictions.
TTT demonstrates that self-supervised updates at test time can improve generalization under distribution shifts \cite{Sun2019TTT}.
TTT-MAE strengthens one-sample adaptation by using reconstruction objectives based on masked autoencoders as the self-supervised signal \cite{Gandelsman2022TTTMAE}.
MEMO adapts at test time by leveraging multiple augmentations and reducing the entropy of the predictive distribution \cite{Zhang2021MEMO}.
A limitation of this family is the additional optimization and augmentation cost incurred at inference, which motivates careful control of update scope and computational overhead in edge or real-time deployments.

\subsection{Non-Global Updates and Source-Free Settings}
\label{subsec:other_adapt}
Beyond full gradient-based updates, another line of work adjusts classifiers or representations at test time.
T3A proposes a lightweight, model-agnostic procedure to adjust the classifier using templates/prototypes under domain shifts \cite{Iwasawa2021T3A}.
SHOT studies a source-free setting where only target data are available, combining information maximization with pseudo-labeling to adapt the feature extractor while transferring the source hypothesis \cite{Liang2020SHOT}.
Although these methods differ in the update target and mechanism, they commonly keep the model's active computational paths fixed and focus on adapting representations and/or classifiers.

\subsection{Inference-Time Adaptation for Language/Foundation Models and PEFT}
\label{subsec:peft}
Inference-time adaptation has also been explored for language models and foundation models.
Dynamic Evaluation continuously updates model parameters at test time to better match recent context, improving sequence prediction performance \cite{Krause2018DynamicEval}.
More recently, parameter-efficient test-time adaptation has gained traction, where only small modules such as prompts, adapters, or LoRA parameters are updated instead of the full model \cite{Zhu2024SelfTPT, Yoon2024CTPT, Imam2024TTL}.
While this trend makes ``partial learning'' more practical by reducing the number of updated parameters, it typically targets adaptation efficiency primarily through parameter count, and is conceptually distinct from approaches that \emph{dynamically reduce inference computation} via input-dependent activation paths.

\subsection{Our Position: Combining Structural Plasticity with Partial Learning}
\label{subsec:our_position}
This work follows the test-time ``partial update'' philosophy of prior TTA/TTT methods, but differs in that adaptation is not confined to normalization, prompts, or lightweight adapters.
Instead, we introduce \emph{input-dependent gating (routing)} to determine active paths and enable \emph{structural sparse execution (conditional computation)}.
Moreover, we explicitly separate online updates into \emph{gate-parameter updates} (\emph{$\theta$-only}), \emph{partial path-weight updates} (\emph{W-only}), or \emph{both}, allowing systematic control of cost and stability.
We further quantify structural fluctuations during adaptation using routing-variability metrics (e.g., flip-rate), and analyze the trade-off among stability, accuracy, and efficiency.
Overall, we extend ``inference-time partial learning'' beyond parameter fine-tuning to an edge/streaming-friendly framework that integrates \emph{structural (path-level) plasticity}.

\section{Why conventional models struggle to train and infer simultaneously}
Consider a standard dense neural network:
\begin{equation}
\hat{y}_t = f(x_t; W_t),
\end{equation}
where the stochastic gradient descent (SGD) update is~\cite{robbins1951stochastic,bottou2010sgd,kingma2015adam}:
\begin{equation}
W_{t+1} = W_t - \eta \nabla_W \mathcal{L}\big(f(x_t; W_t), y_t\big).
\end{equation}

During inference, one usually assumes that $f(\cdot;W)$ is \textbf{fixed}~\cite{bottou2010sgd}. But if $W_t$ changes during inference, then the mapping $x \mapsto \hat{y}$ itself changes over time~\cite{shalevshwartz2012online,zinkevich2003online}. As a result, the same input can produce different outputs, and the inference function is no longer well-defined in the traditional sense~\cite{shalevshwartz2012online}.

Therefore, in dense models, learning--inference concurrency is not merely an implementation inconvenience but ultimately a problem of \textbf{mathematical instability of definition}~\cite{zinkevich2003online,shalevshwartz2012online}.

\section{Why DynamicGate-MLP can perform inference while learning simultaneously}
\begin{figure}
    \centering
    \includegraphics[width=0.75\linewidth]{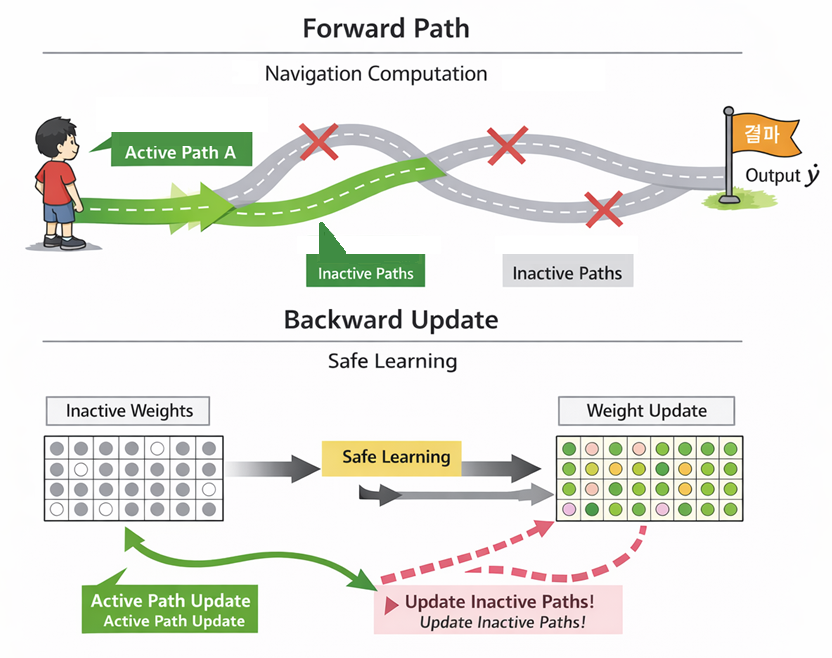}
    \caption{Conceptual diagram of concurrent learning during inference}
    \label{fig:placeholder}
\end{figure}
\begin{itemize}
    \item This model solves a task by \textbf{selecting and using only the necessary computation paths}. (green path)~\cite{bengio2015conditional,shazeer2017moe,fedus2022switch}
    \item Unused paths are \textbf{not computed, making it faster and more stable}. (gray paths with X marks)~\cite{bengio2015conditional,fedus2022switch}
    \item While producing an answer, the model can \textbf{quietly update the weights of unused paths in the background}.~\cite{han2015pruning,evci2020rigl,mocanu2018set}
    \item Thus, the current output remains stable while the model \textbf{simultaneously prepares to solve the next problem better}.~\cite{parisi2019lifelong,kirkpatrick2017ewc}
    \item In short, this model has a \textbf{structure that can perform inference (solving) and learning (improving) at the same time}.~\cite{bengio2015conditional,shazeer2017moe}
\end{itemize}

\section{DynamicGate-MLP: Model Equations and Mathematical Basis}

\subsection{Model definition}
DynamicGate-MLP decomposes computation into \textbf{routing} and \textbf{representation}~\cite{bengio2015conditional,shazeer2017moe}. Let input $x \in \mathbb{R}^n$, weight matrix $W \in \mathbb{R}^{d \times h}$, and gate-network parameters $\theta$. The gate function generates an input-dependent mask~\cite{bengio2015conditional,shazeer2017moe}:
\begin{equation}
m(x; \theta) \in \{0,1\}^{d} \quad \text{(or } [0,1]^d\text{)}.
\end{equation}
The forward computation is defined as~\cite{bengio2015conditional,shazeer2017moe}:
\begin{equation}
\hat{y} = f(x; W, \theta) = F\!\Big(x;\, W \odot m(x;\theta)\Big),
\end{equation}
where $\odot$ denotes element-wise multiplication (or an equivalent structural masking operation)~\cite{han2015pruning,han2016deepcompression}.

\subsection{Parameter decomposition}
We divide parameters into two sets~\cite{bengio2015conditional,shazeer2017moe}:
\begin{itemize}
  \item \textbf{Routing (=gating) parameters:} $\theta$,
  \item \textbf{Representation parameters:} $W$.
\end{itemize}
Importantly, not every element of $W$ is always used for every input~\cite{bengio2015conditional}. If we define the active index set as
\begin{equation}
S(x) = \{ i \mid m_i(x;\theta) = 1 \}
\end{equation}
then only the subset corresponding to $i \in S(x)$ contributes to the output computation~\cite{bengio2015conditional,shazeer2017moe}.

\subsection{Sufficient conditions for learning--inference concurrency}
First, define ``well-defined inference'' as follows~\cite{shalevshwartz2012online,zinkevich2003online}.

\begin{definition}[Well-defined inference]
Inference at time $t$ is said to be well-defined if there exists a parameter state $(W_t,\theta_t)$ such that
\begin{equation}
\hat{y}_t = f(x_t; W_t, \theta_t)
\end{equation}
holds~\cite{shalevshwartz2012online}.
\end{definition}

\begin{proposition}[Concurrency holds under gate-only adaptation]
If $W_t$ is fixed during inference and learning updates only $\theta$, then learning and inference can proceed concurrently without violating well-defined inference~\cite{shalevshwartz2012online,zinkevich2003online}.
\end{proposition}

\subsection{Proof (gate-only adaptation)}
Assume:
\begin{equation}
W_{t+1} = W_t,
\end{equation}
\begin{equation}
\theta_{t+1} = \theta_t - \eta \nabla_\theta \mathcal{L}(W_t,\theta_t),
\end{equation}
where $\mathcal{L}(W,\theta)$ may be the form (expected risk + regularization term)~\cite{robbins1951stochastic,bottou2010sgd}. For example,
\begin{equation}
\mathcal{L}(W,\theta)=\mathbb{E}_{(x,y)}\big[\ell(f(x;W,\theta),y)\big]+\lambda\,\Omega(m(x;\theta)).
\end{equation}

The inference output at time $t$ is
\begin{equation}
\hat{y}_t = f(x_t; W_t, \theta_t)
\end{equation}
and the update of $\theta$ is performed \textbf{after} this output is computed~\cite{shalevshwartz2012online}. Therefore, $\hat{y}_t$ is an exact forward computation of the fixed model snapshot $(W_t,\theta_t)$~\cite{shalevshwartz2012online,zinkevich2003online}. For the next input $x_{t+1}$, $(W_t,\theta_{t+1})$ serves as a new snapshot that again defines a valid model~\cite{shalevshwartz2012online}. Thus, at every time step, inference is always well-defined by some snapshot model~\cite{shalevshwartz2012online}. 

\subsection{Selective weight updates on the active subspace}
DynamicGate-MLP permits an even stronger form of concurrency~\cite{mocanu2018set,evci2020rigl}. Suppose we selectively update only the active subspace~\cite{han2015pruning,evci2020rigl}:
\begin{equation}
W_{t+1}^{(i)}=
\begin{cases}
W_t^{(i)}-\eta\,\nabla_{W^{(i)}} \ell\big(f(x_t;W_t,\theta_t),y_t\big), & i\in S(x_t),\\
W_t^{(i)}, & i\notin S(x_t).
\end{cases}
\end{equation}
Observe that if $i \notin S(x_t)$, then $m_i(x_t;\theta_t)=0$, so $W^{(i)}$ does not contribute to $f(x_t;W_t,\theta_t)$~\cite{bengio2015conditional,shazeer2017moe}. Therefore, even if the update is restricted to $i \in S(x_t)$, the definition of the inference output at time $t$ is not corrupted~\cite{han2015pruning,evci2020rigl}.

\begin{proposition}[Concurrency also holds under active-subspace updates]
If updates are restricted to the active subspace induced by the gate, then learning and inference can proceed concurrently while preserving well-defined inference~\cite{mocanu2018set,evci2020rigl}.
\end{proposition}

\subsection{Interpretation: piecewise-stationary model}
DynamicGate-MLP forms a sequence of snapshots over time~\cite{shalevshwartz2012online,li2014parameterserver}:
\begin{equation}
(W_t,\theta_t) \rightarrow (W_{t+1},\theta_{t+1}).
\end{equation}
Inference at each time can be interpreted as a stationary model at the corresponding snapshot~\cite{shalevshwartz2012online}. This is fundamentally different from dense models in which all parameters affect all inputs~\cite{bengio2015conditional,shazeer2017moe}.

\subsection{Asynchronous and practical concurrency}
In real systems, parameter copies for serving and training can be separated~\cite{dean2012distbelief,li2014parameterserver,niu2011hogwild}:
\begin{itemize}
  \item \textbf{Serving parameters} $(W^s,\theta^s)$,
  \item \textbf{Training parameters} $(W^t,\theta^t)$.
\end{itemize}
Inference proceeds as
\begin{equation}
\hat{y} = f(x; W^s,\theta^s),
\end{equation}
and learning proceeds as
\begin{equation}
(W^t,\theta^t)\leftarrow (W^t,\theta^t)-\eta\nabla\mathcal{L}(W^t,\theta^t)
\end{equation}
~\cite{dean2012distbelief,li2014parameterserver,niu2011hogwild}. Then, with periodic synchronization (snapshot),
\begin{equation}
(W^s,\theta^s)\leftarrow \mathrm{Snapshot}(W^t,\theta^t)
\end{equation}
the serving side always operates stably with fixed snapshot parameters~\cite{li2014parameterserver}. Since DynamicGate-MLP often updates only a subset of parameters that are meaningfully involved per input, synchronization cost can be reduced~\cite{bengio2015conditional,shazeer2017moe,li2014parameterserver}. 


\section{Experiments}

\subsection{Experimental Protocol}

This experiment quantitatively compares how online adaptation under distribution shift (data drift) affects performance, stability, and computational cost~\cite{quinonero2009datasetshift,gama2014conceptdrift}. After training each model offline on a clean distribution, we evaluate on a drifted data stream by (i) measuring accuracy before adaptation (DriftBefore), and (ii) performing online adaptation and then measuring drifted-distribution accuracy after adaptation (AdaptAcc) and clean-distribution performance change (CleanDrop)~\cite{quinonero2009datasetshift,wang2021tent,sun2020ttt}. We also record the degree of prediction changes before/after adaptation (Flip), activation ratio (AR), and relative FLOPs (FLOPs\textasciitilde{}) to jointly analyze efficiency and stability~\cite{wang2021tent,kirkpatrick2017ewc}.

\subsection{Compared Models}

\begin{itemize}
    \item Dense: A standard MLP without sparse routing (gate/router). Since it has no $\theta$ or inactive-W concept required by online adaptation modes (A/B/C/D), those modes are skipped~\cite{bottou2010sgd}.

    \item DG-Hard: Hard-gating-based DynamicGate. It is efficient with low AR/FLOPs, but due to the non-differentiability of hard selection, $\theta$-only (B) updates may be implementation-limited~\cite{bengio2015conditional,fedus2022switch}.

    \item DG-Soft: Soft-gating-based DynamicGate. Flexible routing yields large drift recovery (Recovery), but may increase Flip and CleanDrop~\cite{wang2021tent,kirkpatrick2017ewc}.

    \item DG-Anneal: A DG variant that gradually sharpens routing from soft to hard (annealing). It aims for a performance--stability compromise~\cite{fedus2022switch}.

    \item MoE-Top1: Top-1 hard-routing MoE. It is efficient with low AR/FLOPs, and in this setup online adaptation is performed only in mode D~\cite{shazeer2017moe,fedus2022switch}.

    \item MoE-Soft: Soft-routing MoE. All experts are partially active (AR$\approx$1), so computational cost is high, but final AdaptAcc tends to be high~\cite{shazeer2017moe,lepikhin2021gshard,du2022glam}.
\end{itemize}

\subsection{Online Adaptation Modes (A/B/C/D)}

\begin{itemize}
    \item A (None): No parameter update during the online stage (baseline)~\cite{shalevshwartz2012online}.

    \item B ($\theta$-only): Update only routing/gating parameters $\theta$, with W frozen. Fast, but may increase Flip and risk CleanDrop (forgetting)~\cite{wang2021tent,kirkpatrick2017ewc}.

    \item C (Inactive-W only): Update only representation weights W with low contribution (inactive) for the current input, with $\theta$ frozen. Aims for stability and clean-performance preservation~\cite{kirkpatrick2017ewc,parisi2019lifelong}.

    \item D ($\theta$ + Inactive-W): Update $\theta$ and inactive W together. It generally provides the highest AdaptAcc, but stability depends on the model/routing type~\cite{wang2021tent,parisi2019lifelong}.
\end{itemize}

\subsection{Evaluation Metrics}

\begin{itemize}
    \item DriftBefore: Drifted-distribution accuracy before online adaptation (\%)~\cite{quinonero2009datasetshift,gama2014conceptdrift}.

    \item AdaptAcc: Drifted-distribution accuracy after online adaptation (\%)~\cite{wang2021tent,sun2020ttt}.

    \item Recovery: AdaptAcc - DriftBefore (\%p). Accuracy gain recovered under drift~\cite{gama2014conceptdrift}.

    \item CleanDrop: Change in clean-distribution accuracy after adaptation (\%p). Closer to 0 indicates less forgetting~\cite{kirkpatrick2017ewc,delange2019survey}.

    \item Flip: Ratio of samples whose predicted class changes before/after adaptation (0\textasciitilde{}1). Larger values may indicate instability/forgetting~\cite{wang2021tent,kirkpatrick2017ewc}.

    \item AR: Average activation ratio (ratio activated by gate/router)~\cite{bengio2015conditional,shazeer2017moe,fedus2022switch}.

    \item FLOPs: Relative computation cost (1.00 = full computation)~\cite{bengio2015conditional,fedus2022switch,han2016deepcompression}.

    \item $\theta$/W params: Number of parameters updated during online adaptation~\cite{bengio2015conditional,shazeer2017moe}.
\end{itemize}

\subsection{Results}

Table 1 summarizes performance by model and online adaptation mode~\cite{quinonero2009datasetshift,gama2014conceptdrift}.

\begin{table*}[h]
\centering
\scriptsize
\setlength{\tabcolsep}{3pt}
\renewcommand{\arraystretch}{1.1}
\caption{Summary of online adaptation results under drift}
\label{tab:drift_adapt}
\resizebox{\textwidth}{!}{%
\begin{tabular}{l l c c c c c c c c l}
\toprule
Model & Mode & DriftBefore & AdaptAcc & Recovery & CleanDrop & Flip & AR & FLOPs\textasciitilde{} & $\theta$/W params & Status \\
\midrule
Dense & A\_none & 36.42 & - & - & - & - & - & - & 0 & SKIP(no trainable params) \\
Dense & B\_theta\_only & 36.42 & - & - & - & - & - & - & 0 & SKIP(mode=theta\_only) \\
Dense & C\_w\_inactive\_only & 36.42 & - & - & - & - & - & - & 0 & SKIP(mode=w\_inactive\_only) \\
Dense & D\_theta\_and\_w\_inactive & 36.42 & - & - & - & - & - & - & 0 & SKIP(mode=theta\_and\_w\_inactive) \\
DG-Hard & A\_none & 43.34 & - & - & - & - & - & - & 0 & SKIP(no trainable params) \\
DG-Hard & B\_theta\_only & 43.34 & - & - & - & - & - & - & 0 & SKIP(mode=theta\_only) \\
DG-Hard & C\_w\_inactive\_only & 43.34 & 74.72 & 31.38 & -5.14 & 0.085 & 0.297 & 0.30 & 203,530 & OK \\
DG-Hard & D\_theta\_and\_w\_inactive & 43.34 & 76.24 & 32.90 & -5.47 & 0.070 & 0.297 & 0.30 & 269,322 & OK \\
DG-Soft & A\_none & 27.65 & - & - & - & - & - & - & 0 & SKIP(no trainable params) \\
DG-Soft & B\_theta\_only & 27.65 & 78.25 & 50.60 & -17.30 & 0.982 & 0.679 & 0.68 & 65,792 & OK \\
DG-Soft & C\_w\_inactive\_only & 27.65 & 80.72 & 53.07 & -15.65 & 0.982 & 0.686 & 0.69 & 203,530 & OK \\
DG-Soft & D\_theta\_and\_w\_inactive & 27.65 & 83.73 & 56.08 & -8.05 & 0.945 & 0.743 & 0.74 & 269,322 & OK \\
DG-Anneal & A\_none & 29.24 & - & - & - & - & - & - & 0 & SKIP(no trainable params) \\
DG-Anneal & B\_theta\_only & 29.24 & 72.88 & 43.64 & -16.91 & 0.714 & 0.673 & 0.67 & 65,792 & OK \\
DG-Anneal & C\_w\_inactive\_only & 29.24 & 80.21 & 50.97 & -12.88 & 0.694 & 0.699 & 0.70 & 203,530 & OK \\
DG-Anneal & D\_theta\_and\_w\_inactive & 29.24 & 83.61 & 54.37 & -5.83 & 0.702 & 0.779 & 0.78 & 269,322 & OK \\
MoE-Top1 & A\_none & 31.04 & - & - & - & - & - & - & 0 & SKIP(no trainable params) \\
MoE-Top1 & B\_theta\_only & 31.04 & - & - & - & - & - & - & 0 & SKIP(mode=theta\_only) \\
MoE-Top1 & C\_w\_inactive\_only & 31.04 & - & - & - & - & - & - & 0 & SKIP(mode=w\_inactive\_only) \\
MoE-Top1 & D\_theta\_and\_w\_inactive & 31.04 & 83.66 & 52.62 & -5.74 & 0.490 & 0.250 & 0.25 & 529,934 & OK \\
MoE-Soft & A\_none & 45.52 & - & - & - & - & - & - & 0 & SKIP(no trainable params) \\
MoE-Soft & B\_theta\_only & 45.52 & 50.99 & 5.47 & -0.33 & 0.447 & 1.000 & 1.00 & 1,028 & OK \\
MoE-Soft & C\_w\_inactive\_only & 45.52 & - & - & - & - & - & - & 0 & SKIP(mode=w\_inactive\_only) \\
MoE-Soft & D\_theta\_and\_w\_inactive & 45.52 & 84.60 & 39.08 & -8.15 & 0.491 & 1.000 & 1.00 & 529,934 & OK \\
\bottomrule
\end{tabular}%
}
\end{table*}

\noindent The main observations are as follows~\cite{quinonero2009datasetshift,gama2014conceptdrift}.

\begin{itemize}
    \item \noindent Dense has no routing ($\theta$) or inactive-W concept, so all online adaptation modes were skipped~\cite{bottou2010sgd}.

    \item \noindent For most sparse-routing models, D ($\theta$+inactive W) achieved the highest AdaptAcc~\cite{wang2021tent}. MoE-Soft+D recorded the best AdaptAcc at 84.60\%~\cite{shazeer2017moe,lepikhin2021gshard,du2022glam}.

    \item \noindent From the computation-efficiency perspective, MoE-Top1+D achieved AdaptAcc 83.66\% at FLOPs\textasciitilde{} 0.25, showing the best performance-cost tradeoff~\cite{fedus2022switch}.

    \item \noindent In terms of stability (Flip, CleanDrop), DG-Soft showed high Recovery (+50\textasciitilde{}56\%p) but also large side effects with Flip$\approx$0.95\textasciitilde{}0.98 and CleanDrop up to -17\%p~\cite{wang2021tent,kirkpatrick2017ewc,delange2019survey}.

    \item \noindent DG-Anneal improved stability/clean retention over DG-Soft in mode D, with AdaptAcc 83.61\%, CleanDrop -5.83\%p, and Flip 0.702~\cite{kirkpatrick2017ewc,delange2019survey}.

    \item \noindent DG-Hard was very stable and lightweight with Flip 0.07\textasciitilde{}0.085 and FLOPs\textasciitilde{} 0.30, but AdaptAcc remained in the 74\textasciitilde{}76\% range, lower than the top models~\cite{fedus2022switch,han2016deepcompression}.
\end{itemize}

\subsection{Summary of Experimental Results}

The results show that (i) joint adaptation of routing ($\theta$) and representation weights (W) is effective for drift recovery, and (ii) soft routing provides high adaptation performance but can accompany prediction flips (Flip) and degradation on clean data (CleanDrop)~\cite{wang2021tent,sun2020ttt,kirkpatrick2017ewc}. Therefore, in real-time/edge scenarios, hard-routing (Top1/Hard)-based low-cost adaptation may be advantageous, whereas when soft routing is used, stabilization mechanisms (e.g., routing-change regularization, conservative $\theta$ updates, annealing) are needed to control Flip and CleanDrop~\cite{fedus2022switch,shazeer2017moe}.


\section{Experimental Results by Dataset}
\label{sec:experiments}

\subsection{Summary: Setup and Metrics}
In this section, we compare for each model (i) clean-distribution accuracy (CleanAcc), (ii) drifted-distribution accuracy (DriftAcc), and (iii) post-online-adaptation accuracy (AdaptAcc) under distribution shift (drift).
We also report the prediction flip ratio in the drift segment (Flip(drift)), mean activation ratio (AR(drift)), relative computation (FLOPs\textasciitilde{}\footnote{FLOPs\textasciitilde{} is a FLOPs proxy, i.e., normalized relative computation (Compute Proxy) with respect to a reference model (e.g., Dense) set to 1.00. Lower values indicate lower computational cost.}), and the number of routing parameters ($\theta$ params),
to quantitatively analyze the tradeoff among performance, stability, and efficiency. The performance metrics are as follows.
\begin{itemize}
    \item CleanAcc (clean accuracy): Accuracy of the model on data without drift.
    \item DriftAcc (drift accuracy): Accuracy of the model on data with drift. This indicates how vulnerable the model is to drift.
    \item AdaptAcc (accuracy after adaptation): Accuracy on drifted data after online adaptation of only the router/gate parameters. This shows the effect of online adaptation.
    \item Flip(drift) (drift flip rate): How often routing decisions change under drift (lower is more stable). Dense has no routing, so it is marked as `None`.
    \item AR(drift) (drift activation ratio): The ratio of active units or experts used under drift. This is a proxy for computational cost.
    \item FLOPs~ (FLOPs proxy): An approximation of computation cost proportional to activation ratio (lower is cheaper).
    \item θ params (θ parameters): The number of gate/router parameters trained online.
\end{itemize}

\subsection{Quantitative Results (Final Summary Table)}
Table~\ref{tab:final_summary} summarizes the final results by model.

\begin{table}[h]
\centering
\small
\setlength{\tabcolsep}{6pt}
\renewcommand{\arraystretch}{1.15}
\caption{Table of performance metrics for each model}
\label{tab:final_summary}
\begin{tabular}{l|r r r r r r r}
\toprule
Model & CleanAcc & DriftAcc & AdaptAcc & Flip(drift) & AR(drift) & FLOPs\textasciitilde{} & $\theta$ params \\
\midrule
Dense     & 96.17 & 28.48 & 28.56 & None  & 1.000 & 1.00 & 0 \\
DG-Hard   & 94.80 & 33.97 & 34.21 & 0.104 & 0.297 & 0.30 & 65792 \\
DG-Soft   & 96.32 & 22.42 & 74.00 & 0.953 & 0.908 & 0.91 & 65792 \\
DG-Anneal & 95.58 & 26.67 & 61.60 & 0.996 & 0.913 & 0.91 & 65792 \\
MoE-Top1  & 96.85 & 36.74 & 37.48 & 0.615 & 0.250 & 0.25 & 1028 \\
MoE-Soft  & 96.90 & 39.67 & 44.68 & 0.429 & 1.000 & 1.00 & 1028 \\
\bottomrule
\end{tabular}
\end{table}

\subsection{Graph Descriptions}
\paragraph{Overview} The figures below are intended to visually compare the experimental results and are organized to show:
(1) Online Adaptation Loss ($\theta$-only), (2) Accuracy under Drift: Before vs After $\theta$-only Online Adaptation,
(3) Compute Proxy (FLOPs\textasciitilde{}) under Drift (lower is cheaper),
and (4) Routing Flip Rate under Drift (higher = less stable), respectively.

\begin{figure}[!t]
  \centering
  \IfFileExists{figures/MNIST_FIG1_OnlineAdaptionLoss.png}{%
    \includegraphics[width=0.95\linewidth]{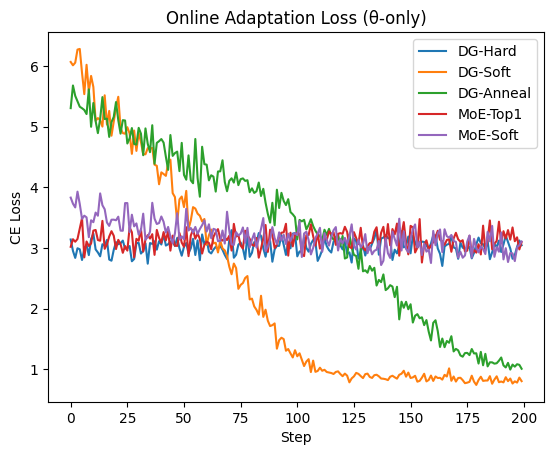}%
  }{%
    \fbox{\parbox[c][0.28\textheight][c]{0.95\linewidth}{\centering
    (blank) figures/MNIST\_FIG1\_OnlineAdaptionLoss.png}}%
  }
  
  \caption{(1) Online Adaptation Loss ($\theta$-only)\\
  A graph showing the loss change during online adaptation steps for each gating/MoE model. The Dense model has no trainable gate/router parameters, so it has no loss curve.}
  \label{fig:mnist_fig1_loss}
\end{figure}
\paragraph{Explanation} As Step progresses beyond 100, the green DG-Anneal curve gradually decreases, while the orange DG-Soft curve drops sharply around Step 100. As a result, both converge near Step 200. This indicates that the loss clearly decreases after online adaptation, and correspondingly accuracy increases. DG-Anneal improves substantially from DriftAcc 26.67\% to 61.60\% after adaptation, and DG-Soft also improves substantially from DriftAcc 22.42\% to 74.0\% after adaptation.

\begin{figure}[!t]
  \centering
  \IfFileExists{figures/MNIST_FIG2_AccuracyunderDrift.png}{%
    \includegraphics[width=0.95\linewidth]{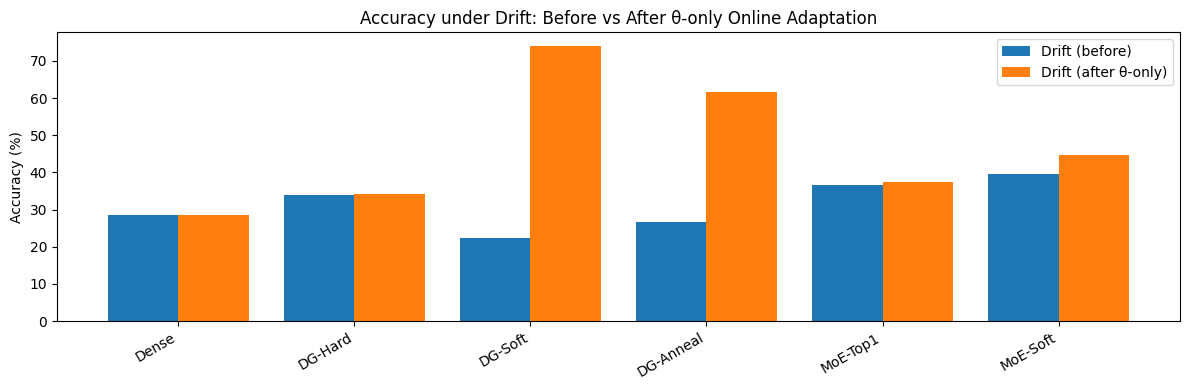}%
  }{%
    \fbox{\parbox[c][0.28\textheight][c]{0.95\linewidth}{\centering
    (blank) figures/MNIST\_FIG2\_AccuracyunderDrift.png}}%
  }
  \caption{(2) Accuracy under Drift\\Before vs After $\theta$-only Online Adaptation: A bar chart comparing how each model's accuracy changes before and after online adaptation in a drift environment.}
  \label{fig:mnist_fig2_acc_drift}
\end{figure}

\noindent As shown in Figure 2 above, DG-Soft and DG-Anneal visually demonstrate that under drift, accuracy after online adaptation increases substantially after only the gating/routing values are changed, consistent with the explanation of Figure 1. 

\begin{figure}[!t]
  \centering
  \IfFileExists{figures/MNIST_FIG3_ComputeProxy.png}{%
    \includegraphics[width=0.95\linewidth]{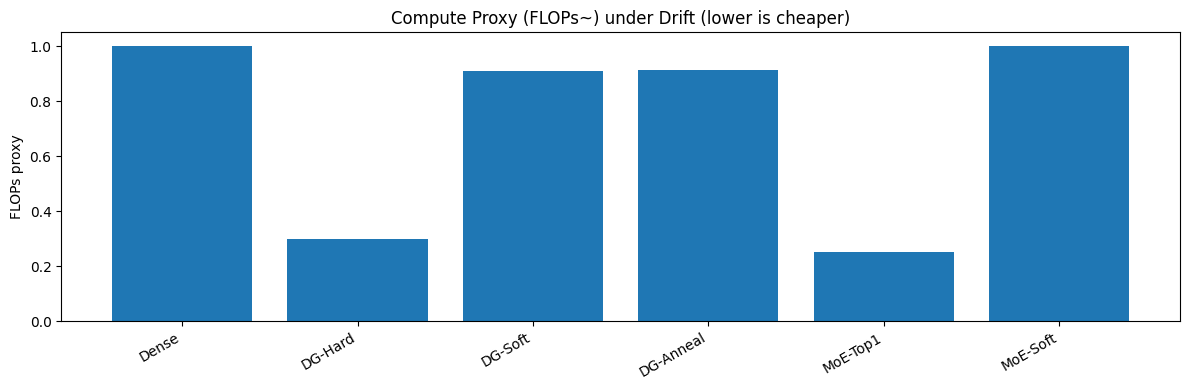}%
  }{%
    \fbox{\parbox[c][0.28\textheight][c]{0.95\linewidth}{\centering
    (blank) figures/MNIST\_FIG3\_ComputeProxy.png}}%
  }
  \caption{(3) Compute Proxy (FLOPs\textasciitilde{}) under Drift (lower is cheaper)\\ A bar chart showing the computational cost proxy of each model in a drift environment. Corresponds to AR(drift).}
  \label{fig:mnist_fig3_compute}
\end{figure}
\noindent The computation-reduction metric shows that DG-Soft and DG-Anneal provide some reduction compared with Dense and MoE-Soft. 

\begin{figure}[!t]
  \centering
  \IfFileExists{figures/MNIST_FIG4_Routing_FlipRate.png}{%
    \includegraphics[width=0.95\linewidth]{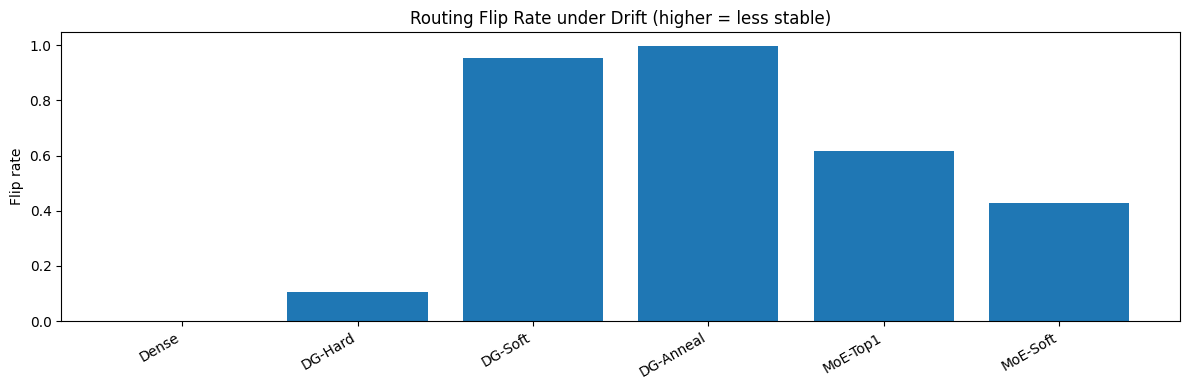}%
  }{%
    \fbox{\parbox[c][0.28\textheight][c]{0.95\linewidth}{\centering
    (blank) figures/MNIST\_FIG4\_Routing\_FlipRate.png}}%
  }
  \caption{(4) Routing Flip Rate under Drift (higher = less stable)\\ A bar chart showing the stability of routing decisions in a drift environment. Higher values mean routing decisions change more frequently.}
  \label{fig:mnist_fig4_fliprate}
\end{figure}

\noindent In the routing Flip rate under drift, values are relatively high at 0.953--0.996 compared with other models. However, a high Flip rate is not always bad. Flip indicates how often routing decisions change under drift, and in models with flexible gating mechanisms, frequent routing changes can be an important part of successful adaptation to environmental changes. Please refer to the later section on correlation analysis between Flip ratio and AdaptAcc. 

\noindent In conclusion, this experiment shows that online adaptation of gate parameters under data drift can substantially improve performance, especially in models such as DG-Soft and DG-Anneal. In the DG-Hard model, due to the hard-gating property, gradient flow is interrupted and the model does not show as dynamic adaptation as DG-Soft or DG-Anneal; however, as noted in the DynamicGateLayer comments, this can be improved by adding STE (Straight-Through Estimator).

\FloatBarrier

\subsection{Interpretation and Discussion}
The results in Table~\ref{tab:final_summary} show the following characteristics from the perspectives of \textbf{(i) vulnerability to drift}, \textbf{(ii) online adaptation gain}, \textbf{(iii) stability (Flip)}, and \textbf{(iv) efficiency (FLOPs\textasciitilde{}, AR)}.

\paragraph{(1) Vulnerability to drift: sharp drop in DriftAcc}
All models have high CleanAcc around 94.8--96.9, but DriftAcc drops sharply to 22.4--39.7.
This means that high performance on the clean distribution does not directly translate into robustness under distribution shift,
and suggests that drift handling (adaptation or robustness enhancement) is necessary in real streaming/field environments.

\paragraph{(2) Online adaptation gain: degree of recovery in AdaptAcc}
The recovery in post-adaptation performance (AdaptAcc) varies greatly by model.
\begin{itemize}
  \item Dense changes only marginally from DriftAcc 28.48 to AdaptAcc 28.56.
        This shows that without routing/conditional computation, the ``adaptation'' setting in this experiment does not translate into drift recovery.
  \item DG-Hard rises slightly from DriftAcc 33.97 to AdaptAcc 34.21.
        In other words, even with an efficient structure, the adaptation gain may be limited.
  \item DG-Soft rises sharply from DriftAcc 22.42 to AdaptAcc 74.00.
        Under the same $\theta$ params (65792), the adaptation effect is very large,
        and drift recovery (increase in AdaptAcc) is the most pronounced among the experimental groups.
  \item DG-Anneal rises from DriftAcc 26.67 to AdaptAcc 61.60,
        showing a large recovery, though smaller than DG-Soft.
  \item MoE-Top1 rises slightly from DriftAcc 36.74 to AdaptAcc 37.48.
  \item MoE-Soft rises from DriftAcc 39.67 to AdaptAcc 44.68,
        showing relatively larger adaptation gain within the MoE family.
\end{itemize}
In summary, this experiment shows that \textbf{DG-Soft and DG-Anneal achieve very large recovery under drift}.

\paragraph{(3) Stability (Flip(drift)) tradeoff}
Flip(drift) indicates how often predictions flip in the drift segment (a signal of instability/volatility).
\begin{itemize}
  \item DG-Hard has Flip(drift)=0.104, which is very low and thus highly stable.
        In contrast, its adaptation gain (AdaptAcc increase) is limited.
  \item DG-Soft has Flip(drift)=0.953, and DG-Anneal has 0.996, both very high.
        Thus, large AdaptAcc recovery is accompanied by \textbf{very high prediction volatility}.
  \item MoE-Soft has Flip(drift)=0.429, lower than DG-Soft/DG-Anneal,
        and MoE-Top1 is at an intermediate level with 0.615.
\end{itemize}
Therefore, the results show that \textbf{strong adaptation recovery (high AdaptAcc)} can appear together with
\textbf{high prediction flipping (Flip)}, and when reliability/stability is important in deployment, controlling Flip becomes a key design factor.

\paragraph{(4) Relationship between efficiency (AR, FLOPs\textasciitilde{}) and performance}
AR(drift) and FLOPs\textasciitilde{} represent the efficiency of conditional-computation structures.
\begin{itemize}
  \item MoE-Top1 is the most efficient with AR=0.250 and FLOPs\textasciitilde{}=0.25,
        but its AdaptAcc is 37.48, indicating limited recovery.
  \item DG-Hard is also efficient with AR=0.297 and FLOPs\textasciitilde{}=0.30, and has low Flip.
        Thus it is strong in terms of \textbf{efficiency + stability}.
  \item DG-Soft/DG-Anneal have FLOPs\textasciitilde{}=0.91, close to full computation,
        and high AR values of 0.908/0.913.
        They obtain strong AdaptAcc recovery (74.00, 61.60),
        but pay high costs in efficiency (compute) and stability (Flip).
  \item Dense and MoE-Soft have AR=1.000 and FLOPs\textasciitilde{}=1.00 (full computation),
        and with equal or larger computation than DG-Soft/DG-Anneal,
        their AdaptAcc values are 28.56 and 44.68, respectively.
\end{itemize}
Overall, the results suggest that \textbf{(i) high-efficiency (low FLOPs\textasciitilde{}) models may be stable but may have limited adaptation gain},
and \textbf{(ii) strong adaptation recovery can be coupled with high AR and high Flip}.
Thus, in practical applications, the design choice may diverge depending on the objective:
(A) stability-oriented (DG-Hard, MoE-Soft) or (B) adaptation-performance-oriented (DG-Soft, DG-Anneal).

\paragraph{(5) Practical implications}
\begin{itemize}
  \item \textbf{Stability first (e.g., medical/safety/policy):} Structures with low Flip (e.g., DG-Hard) are favorable,
        and adaptation should be applied conservatively (learning rate/frequency limits) or with snapshot/validation stages.
  \item \textbf{Adaptation performance first (e.g., personalization/rapidly changing environments):} Large AdaptAcc recovery in DG-Soft is attractive,
        but Flip(drift) can become very large, so prediction-volatility mitigation (e.g., calibration, regularization, update limits) is needed.
  \item \textbf{Edge efficiency first (battery/latency):} Low-FLOPs\textasciitilde{} models such as MoE-Top1 are advantageous,
        but additional adaptation design is needed for drift recovery.
\end{itemize}

\section{Correlation Analysis Between Flip Ratio and AdaptAcc}

\noindent We analyze the extracted data from Table 3 to identify patterns and relationships between the Flip ratio and AdaptAcc across various models (Dense, DG-Hard, DG-Soft, DG-Anneal, MoE-Top1, MoE-Soft) and adaptation modes (B, C, D). We discuss whether high/low Flip generally leads to high/low AdaptAcc, or whether the relationship is more nuanced.

\begin{table}[h]
\centering
\small
\setlength{\tabcolsep}{6pt}
\renewcommand{\arraystretch}{1.15}
\caption{Flip and AdaptAcc by model/adaptation mode}
\label{tab:flip_adaptacc}
\begin{tabular}{l l r r}
\toprule
Model & Mode & Flip & AdaptAcc \\
\midrule
DG-Hard   & C\_w\_inactive\_only        & 0.106602 & 0.7467 \\
DG-Hard   & D\_theta\_and\_w\_inactive  & 0.106076 & 0.7991 \\
DG-Soft   & B\_theta\_only              & 0.908429 & 0.7353 \\
DG-Soft   & C\_w\_inactive\_only        & 0.925763 & 0.7882 \\
DG-Soft   & D\_theta\_and\_w\_inactive  & 0.878457 & 0.8447 \\
DG-Anneal & B\_theta\_only              & 0.671266 & 0.6180 \\
DG-Anneal & C\_w\_inactive\_only        & 0.659161 & 0.7505 \\
DG-Anneal & D\_theta\_and\_w\_inactive  & 0.662597 & 0.8307 \\
MoE-Top1  & D\_theta\_and\_w\_inactive  & 0.505895 & 0.8218 \\
MoE-Soft  & B\_theta\_only              & 0.482000 & 0.4399 \\
MoE-Soft  & D\_theta\_and\_w\_inactive  & 0.362000 & 0.8392 \\
\bottomrule
\end{tabular}
\end{table}

\begin{figure}[!t]
  \centering
  \IfFileExists{figures/Correlation.png}{%
    \includegraphics[width=0.95\linewidth]{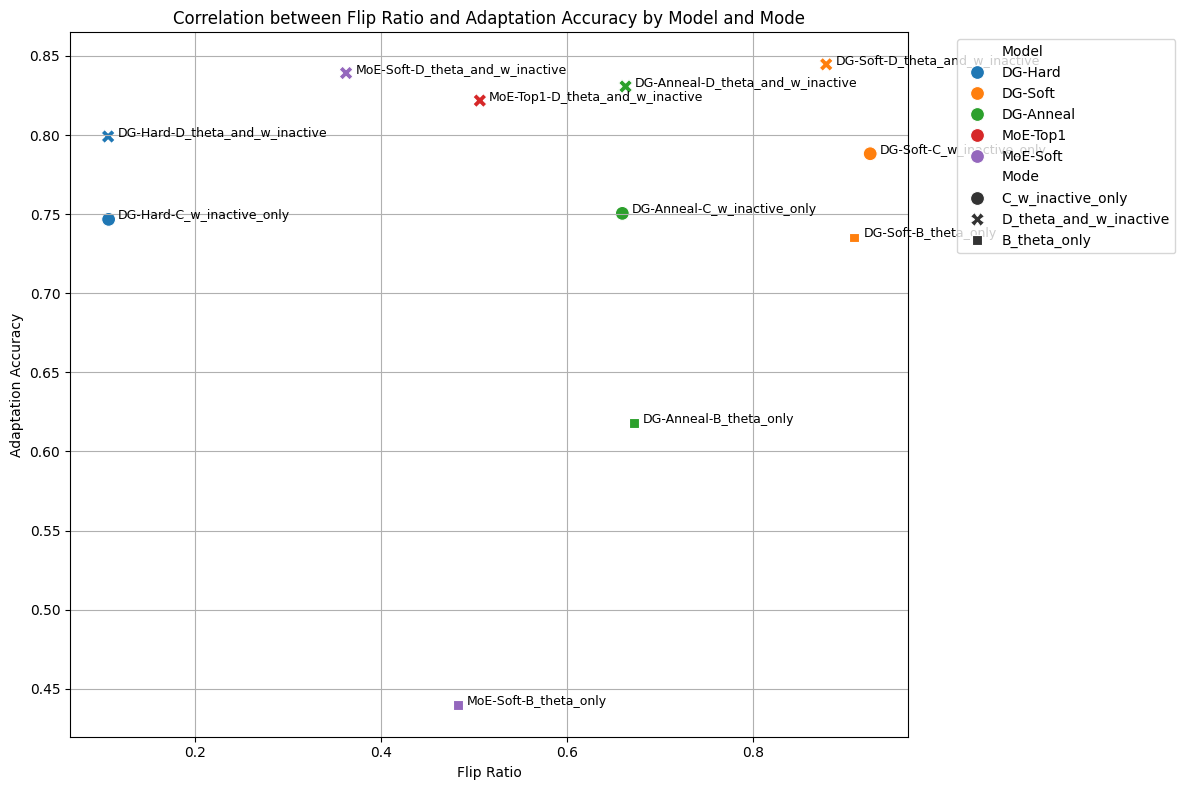}%
  }{%
    \fbox{\parbox[c][0.28\textheight][c]{0.95\linewidth}{\centering
    (blank) figures/Correlation.png}}%
  }
  \caption{Correlation graph between Flip ratio and AdaptAcc}
  \label{fig:correlation}
\end{figure}

\noindent Correlation analysis between \texttt{Flip} ratio and \texttt{AdaptAcc}.
The scatter plot analysis of the relationship between \texttt{Flip} ratio and \texttt{AdaptAcc} reveals several patterns and model-specific characteristics.

\paragraph{(1) Overall relationship}
\begin{itemize}
  \item \textbf{Low Flip with relatively high AdaptAcc:}
  The DG-Hard model's \texttt{C\_ w\_inactive\_only} and \texttt{D\_theta\_and\_w\_inactive} modes show very low \texttt{Flip} values around 0.1, yet relatively high \texttt{AdaptAcc} values of about 0.74 and 0.79, respectively. This suggests that even when routing changes little due to hard gating, meaningful adaptation performance can still be achieved through appropriate update strategies (e.g., inactive W updates or $\theta$+inactive W updates).
  
  \item \textbf{High Flip with high AdaptAcc:}
  In DG-Soft and DG-Anneal \texttt{D\_theta\_and\_w\_inactive} modes, and MoE-Soft \texttt{D\_theta\_and\_w\_inactive}, \texttt{Flip} is relatively high (0.36--0.87), and \texttt{AdaptAcc} is also among the highest (0.83--0.84). This implies that even when routing changes frequently and flexibly, more effective adaptation is possible when routing decisions and related weights are updated together.
  
  \item \textbf{Cases with high Flip but relatively low AdaptAcc:}
  DG-Soft \texttt{B\_theta\_only} (\texttt{Flip} $\approx 0.90$, \texttt{AdaptAcc} $\approx 0.73$) and DG-Anneal \texttt{B\_theta\_only} (\texttt{Flip} $\approx 0.67$, \texttt{AdaptAcc} $\approx 0.61$) have high \texttt{Flip} but relatively low \texttt{AdaptAcc}. This shows that the router-only update strategy (\texttt{theta\_only}) may have limitations in fully exploiting flexible gating (high \texttt{Flip}) to increase performance. In particular, MoE-Soft \texttt{B\_theta\_only} has a mid-level \texttt{Flip} of about 0.48 but a low \texttt{AdaptAcc} of about 0.44, suggesting that updating only the router is not very effective in soft MoE.
\end{itemize}

\paragraph{(2) Model-type-specific characteristics}
\begin{itemize}
  \item \textbf{DG-Hard:} Maintains low \texttt{Flip} while showing decent adaptation performance. It is stable because routing changes little due to hard gating.
  
  \item \textbf{DG-Soft / DG-Anneal:} Tend to show high \texttt{Flip}, and achieve the highest \texttt{AdaptAcc} particularly in \texttt{D\_theta\_and\_w\_inactive}. This suggests that the ability to flexibly change routing is important for responding to environmental changes. In addition, the \texttt{Anneal} scheme leaves room to control routing volatility via a temperature schedule, enabling attempts at a compromise between adaptation performance and stability.
  
  \item \textbf{MoE family:} MoE-Top1 has mid-level \texttt{Flip} around 0.50, while MoE-Soft appears in the range 0.36--0.48. Both MoE models show high \texttt{AdaptAcc} in \texttt{D\_theta\_and\_w\_inactive}, and MoE-Soft in particular is most effective when the router and related weights are updated together.
\end{itemize}

\paragraph{(3) Routing stability vs. adaptation effect}
\begin{itemize}
  \item \textbf{Routing stability (low Flip):} As seen in DG-Hard, this provides predictable behavior and consistency. However, adaptation capability to a new environment is primarily expressed in specific update modes (\texttt{C\_w\_inactive\_only}, \texttt{D\_theta\_and\_w\_inactive}).
  
  \item \textbf{Adaptation effect (high AdaptAcc):} Most pronounced in cases with relatively high \texttt{Flip}, such as DG-Soft, DG-Anneal, and MoE-Soft under \texttt{D\_theta\_and\_w\_inactive}. This suggests that actively reconfiguring internal routing under drift can contribute to performance recovery.
\end{itemize}

\paragraph{Summary}
Rather than a simple linear relationship, \texttt{Flip} and \texttt{AdaptAcc} exhibit a complex relationship determined by the model's gating/routing mechanism and the update mode. High \texttt{Flip} is not always negative; in flexible gating structures, routing change itself can be part of successful adaptation. In particular, the \texttt{D\_theta\_and\_w\_inactive} mode was effective in achieving high \texttt{AdaptAcc} across many gated models, suggesting that simultaneously (or selectively) updating related weights together with routing-decision changes is important for adaptation in drift environments.

\section{Discussion}
\subsection{Why this is not a ``hack''}
The serving output is always defined by some valid parameter snapshot~\cite{shalevshwartz2012online,li2014parameterserver}. In addition, if snapshots are fixed at the serving boundary, partially updated parameters do not get mixed into the output of a single request~\cite{li2014parameterserver,niu2011hogwild}. This is compatible with standard definitions of online learning and deployed inference~\cite{shalevshwartz2012online,zinkevich2003online}.

\subsection{Biological/neuromorphic perspective}
The intuition that synaptic strength changes relatively slowly while active pathways or attention adapt quickly is analogous to the routing--representation decomposition in this structure.

\subsection{Hardware/system perspective}
DynamicGate-MLP naturally aligns with block-sparse accelerators, enables online adaptation without full-model retraining, and provides a practical path toward on-device continual learning~\cite{han2016deepcompression,fedus2022switch}.

\section{Conclusion}
This paper showed that DynamicGate-MLP permits learning--inference concurrency through the structural separation of routing and representation parameters~\cite{bengio2015conditional,shazeer2017moe}. Through sufficient conditions and proofs, we established that even under asynchronous or partial updates, the inference output at each time step can be interpreted as a forward computation of a valid snapshot model~\cite{dean2012distbelief,li2014parameterserver,niu2011hogwild}. This provides a foundation for adaptive neural systems that can overcome the concurrency limitations of conventional dense models~\cite{quinonero2009datasetshift,gama2014conceptdrift,parisi2019lifelong}.

\section{Acknowledgements}

In this work, I implemented the experimental code and produced experimental data through iterative prototyping using multiple generative-AI tools (so-called ``vibe coding). Generative tools were mainly used for code-structure design, repetitive implementation, and debugging assistance. The experimental settings, data-generation procedures, result interpretation, and conclusions were verified and are the responsibility of the author. I note that automatically generated artifacts can contain errors, and all outputs were reviewed and edited before inclusion.

\bibliographystyle{unsrt}
\bibliography{references}

\end{document}